\documentclass[runningheads]{llncs}

\usepackage{eccv}

\usepackage{eccvabbrv}

\usepackage{graphicx}
\usepackage{booktabs}

\usepackage[accsupp]{axessibility}  
\usepackage{soul}

\usepackage[pagebackref,breaklinks,colorlinks,citecolor=eccvblue]{hyperref}

\begin{document}

\title{ParaPoint: Learning Global Free-Boundary Surface Parameterization of 3D Point Clouds}


\author{Qijian Zhang\inst{1} \and
	Junhui Hou\inst{1}*\and
	Ying He\inst{2}}

\authorrunning{Q. Zhang et al.}

\institute{Department of Computer Science, City University of Hong Kong \and
	Nanyang Technological University, Singapore\\ 
	\email{qijizhang3-c@my.cityu.edu.hk;~ jh.hou@cityu.edu.hk;~yhe@ntu.edu.sg}
}

\maketitle

\begin{abstract}
	Surface parameterization is a fundamental geometry processing problem with rich downstream applications. Traditional approaches are designed to operate on well-behaved mesh models with high-quality triangulations that are laboriously produced by specialized 3D modelers, and thus unable to meet the processing demand for the current explosion of ordinary 3D data. In this paper, we seek to perform UV unwrapping on unstructured 3D point clouds. Technically, we propose ParaPoint, an unsupervised neural learning pipeline for achieving global free-boundary surface parameterization by building point-wise mappings between given 3D points and 2D UV coordinates with adaptively deformed boundaries. We ingeniously construct several geometrically meaningful sub-networks with specific functionalities, and assemble them into a bi-directional cycle mapping framework. We also design effective loss functions and auxiliary differential geometric constraints for the optimization of the neural mapping process. To the best of our knowledge, this work makes \textit{the first attempt} to investigate neural point cloud parameterization that pursues both global mappings and free boundaries. Experiments demonstrate the effectiveness and inspiring potential of our proposed learning paradigm. The code will be publicly available.
\end{abstract}

\section{Introduction}
As an essential and long-standing problem in the geometry processing community, surface parameterization intuitively refers to the process of flattening a 3D surface onto a 2D plane, which is typically called the parameter domain. For any 3D spatial point $(x, y, z)$ lying on the underlying shape surface, we explicitly map it to a 2D coordinate $(u, v)$, with some continuity and distortion constraints. Hence, building such a point-to-point mapping is also usually called UV unwrapping, which is widely applied in a rich variety of downstream applications such as texture mapping, remeshing, and editing.

Traditional computation approaches operate on triangular meshes to achieve surface parameterization. Theoretically, for any two surfaces with similar topological structures, there exists a bijective mapping between them. However, when the topology of the target 3D surface becomes complex, we need to pre-open the original mesh along appropriate cutting seams to a developable disk. Therefore, the current industrial practice for UV unwrapping of 3D models consists of two steps, i.e., manually specifying the cutting seams on the given mesh, and then applying classic disk-topology parameterization algorithms, such as LSCM~\cite{levy2023least} and ABF++~\cite{sheffer2005abf++}, which have been well integrated into many mature 3D modeling softwares (e.g., Blender). Nevertheless, the existing surface parameterization algorithms as well as conventional industrial workflows still have many aspects of shortcomings and inconveniences. First of all, these approaches only work for well-behaved meshes with high-quality triangulations, which significantly restrict their applications on ordinary 3D data. For complex 3D models, finding optimal cutting strategies still remains a highly subjective task, despite the existence of heuristic seam generation algorithms \cite{sheffer2002seamster,erickson2002optimally}. Moreover, in practice, it turns out that implementing, configuring, and tuning those complicated traditional mesh parameterization algorithms can be cumbersome and exhausting.

In recent years, there has emerged a family of neural parameterization approaches that learn parameterized representations of 3D geometric shapes through deep neural networks. The earlier works of FoldingNet~\cite{yang2018foldingnet} and AtlasNet~\cite{groueix2018papier} are among the best two representatives, which build point-wise mappings via deforming a pre-defined 2D lattice grid to reconstruct the target 3D surface. RegGeoNet~\cite{zhang2022reggeonet} and Flattening-Net~\cite{zhang2023flattening} achieve fixed-boundary (i.e., a pre-defined square area) structurization of irregular 3D point clouds through geometry image \cite{gu2002geometry} representations. However, these approaches are not real-sense surface parameterization owing to the lack of mapping constraints. DiffSR~\cite{bednarik2020shape} and Nuvo~\cite{srinivasan2023nuvo} propose to locally parameterize the whole 3D surface with stronger constraints on the learned neural mapping process. Their difference lies in that DiffSR aggregates multi-patch parameterizations to reconstruct the original surface geometry, while Nuvo assigns surface points to different charts in a probabilistic manner.

In this paper, we make \textit{the first attempt} to achieve neural surface parameterization on unstructured 3D point clouds, featured by global mapping and free boundary. We propose \textit{ParaPoint}, an unsupervised neural learning pipeline, 
which can build point-to-point mappings between given 3D points and 2D UV coordinates with adaptively deformed boundaries. We ingeniously construct a series of geometrically meaningful sub-networks with specific functionalities such as boundary deforming, surface cutting and stitching, as well as wrapping and unwrapping between 2D and 3D domains. These interpretable sub-networks are further assembled into the complete bi-directional cycle mapping framework. To facilitate the optimization of the neural mapping process, we also design effective loss functions and auxiliary differential geometric constraints. Experiments show that ParaPoint is able to automatically find reasonable 3D cutting seams and 2D UV boundaries, leading to satisfactory parameterization results when dealing with 3D models with different levels of geometric and topological complexities.

Generally, the major advantages of our proposed ParaPoint learning framework can be summarized as follows:
\begin{itemize}
	
	\item Compared with traditional mesh parameterization approaches, ParaPoint is directly applicable to unoriented surface points and liberates from additional efforts on surface cutting. Besides, implementing and running such an end-to-end neural learning framework is easy and straightforward.
	
	\item Different from multi-patch local parameterization \cite{groueix2018papier,bednarik2020shape,srinivasan2023nuvo}, ParaPoint pursues global parameterization, which is a much harder yet more valuable setting.
	
	\item Compared with point cloud geometry images \cite{zhang2022reggeonet,zhang2023flattening} adopting fixed parameter domains, ParaPoint adaptively deforms the UV boundaries and explicitly regularizes the neural mapping functions, thus significantly reducing distortions and discontinuities.
	
\end{itemize}

\section{Related Works} \label{sec:rw}

\subsection{Traditional Mesh Parameterization Approaches} \label{sec:rw-mesh-para}

Mesh parameterization \cite{DBLP:books/sp/05/FloaterH05,sheffer2007mesh} has been extensively investigated in the digital geometry processing and computer graphics communities, driven by its diverse applications such as remeshing, morphing, texture mapping, and compression, to name a few. Early approaches typically formulate the problem as a Laplacian problem, where boundary points are anchored to a predetermined convex 2D curve~\cite{10.1145/218380.218440,DBLP:journals/cagd/Floater97a}, and tackled it through the use of sparse linear system solvers. This linear approach is valued for its simplicity, efficiency, and the guarantee of bijectivity it provides. However, the rigidity of fixing boundary points in the parametric domain frequently leads to significant distortions. In response to these challenges, free-boundary parameterization techniques have been developed~\cite{sheffer2005abf++,DBLP:journals/cgf/LiuZXGG08}, offering a more flexible setting by easing the boundary constraints. While these methods bring certain improvements, they often struggle to maintain global bijectivity. More recent approaches have shifted towards minimizing simpler proxy energies~\cite{DBLP:journals/tog/SmithS15,DBLP:journals/tog/RabinovichPPS17}, adopting a strategy that alternates between local and global optimization steps. This approach not only accelerates the convergence rate but also improves the overall quality of the parameterization results. Despite the advancements, all these methods fundamentally rely on the connectivity inherent in mesh structures, casting uncertainty on their applicability to point clouds. Furthermore, these mesh-oriented parameterization methods typically assume that the input meshes possess relatively regular triangulations. Consequently, when faced with input meshes of inferior quality, characterized by irregular triangulations or anomalies, remeshing becomes an indispensable step.

\subsection{Neural Parameterization Approaches} \label{sec:rw-neural-para}

In recent years, there also exists a line of works applying neural networks for learning parameterized 3D geometric representations. FoldingNet~\cite{yang2018foldingnet} proposes to deform a uniform 2D grid to reconstruct the target 3D point cloud for unsupervised geometric feature learning. AtlasNet~\cite{groueix2018papier} applies multi-grid deformation to learn locally parameterized representations. Subsequently, a series of follow-up researches inherit such a ``folding-style'' parameterization paradigm as pioneered by \cite{yang2018foldingnet,groueix2018papier} and investigate different aspects of modifications. GTIF~\cite{chen2019deep} introduces graph topology inference and filtering mechanisms, empowering the decoder to preserve more representative geometric features in the latent space. EleStruc~\cite{deprelle2019learning} proposes to perform shape reconstruction from learnable 3D elementary structures, rather than a pre-defined 2D lattice. Similarly, TearingNet~\cite{pang2021tearingnet} adaptively breaks the edges of an initial primitive graph for emulating the topology of the target 3D point cloud, which can effectively deal with higher-genus or multi-object inputs. In fact, the ultimate goal of the above line of approaches is to learn expressive shape codewords by means of deformation-driven 3D surface reconstruction. The characteristics of surface parameterization, i.e., the mapping process between 3D surfaces and 2D parameter domains, are barely considered.

In contrast to the extensive research in the fields of deep learning-based 3D geometric reconstruction and feature learning, there only exist a few studies that particularly focus on neural surface parameterization. DiffSR~\cite{bednarik2020shape} adopts the basic multi-patch reconstruction framework \cite{groueix2018papier} and explicitly regularizes multiple differential surface properties. NSM~\cite{morreale2021neural} explores neural encoding of surface maps by overfitting a neural network to an existing UV parameterization pre-computed via standard mesh parameterization algorithms \cite{tutte1963draw,rabinovich2017scalable}. DA-Wand~\cite{liu2023wand} constructs a parameterization-oriented mesh segmentation framework. Around a specified triangle, it learns to select a local sub-region, which is supposed to be sufficiently developable to produce low-distortion parameterization. Inheriting the geometry image (GI) \cite{gu2002geometry} representation paradigm, RegGeoNet~\cite{zhang2022reggeonet} and Flattening-Net~\cite{zhang2023flattening} propose to learn deep regular representations of unstructured 3D point clouds. However, these two approaches lack explicit constraints on the parameterization distortions, and their local-to-global assembling procedures are hard to control. More recently, Nuvo~\cite{srinivasan2023nuvo} proposes a neural UV mapping framework that operates on oriented 3D points sampled from arbitrary 3D representations, liberating from the stringent quality demands of mesh triangulation. This approach assigns the original surface to multiple charts and ignores the packing procedure, thus essentially differing from our targeted global parameterization setting.

\section{Proposed Method} \label{sec:proposed-method}

Essentially, we seek to learn a point-wise mapping between 3D points located on the underlying shape surface and 2D UV coordinates in the parameter domain. Ideally, this mapping is supposed to be bijective and geometrically meaningful, i.e., the 3D-to-2D mapping should approximate the process of unwrapping the 3D shape surface onto the 2D parameter domain; inversely, the 2D-to-3D mapping should approximate the process of wrapping the 2D UV points onto the 3D shape surface. Additionally, under the problem settings of surface parameterization, it is required that the learned neural parameterization should achieve acceptable levels of distortions.

In what follows, we begin with presenting an overall description of our proposed neural point cloud parameterization approach with definitions of involved mathematical notations in Sec.~\ref{sec:overview}. Then, we introduce the network structures of our key components with explanations of their specific functionalities in Sec.~\ref{sec:network-structures}. After that, we introduce how to assemble these sub-networks to a bi-directional cycle mapping workflow in Sec.~\ref{sec:method-bi-cy}. Finally, Sec.~\ref{sec:training-objectives} formulates a series of different loss functions and auxiliary constraints for the optimization of ParaPoint.

\subsection{Method Overview} \label{sec:overview}

ParaPoint is designed as a bi-directional cycle mapping workflow, which directly operates on unstructured 3D point clouds. The overall learning framework consists of several ingeniously designed sub-networks with specific geometrically meaningful functionalities, mimicking the actual physical procedures of parameterizing a 3D surface onto a 2D planar domain.

Formally, given an input 3D point cloud $\mathbf{P} \in \mathbb{R}^{N \times 3}$ containing $N$ unoriented spatial points sampled from the underlying geometric shape surface, we aim at point-wisely parameterizing (i.e., row-wisely mapping) $\mathbf{P}$ onto the planar domain to produce the corresponding 2D UV coordinates $\mathbf{Q} \in \mathbb{R}^{N \times 2}$. However, learning a global parameterization solely from an unstructured point set typically faces great difficulties, as outlined in the following:
\begin{itemize}
	
	\item Using fixed UV domain boundary typically results in significant parameterization distortions. However, it is highly non-trivial to determine the optimal UV boundary for different 3D shapes.	
	
	\item Due to the complexities of different shape topologies, the underlying surface is usually not homomorphic to a 2D plane, thus being highly undevelopable.
	
	\item Due to the lack of connectivity information, the surface manifolds of thin structures become difficult to recognize and parameterize.
	
\end{itemize}

To deal with the above challenges, we design geometrically meaningful sub-networks for interpretable and controllable neural parameterization, which are assembled to build an effective bi-directional cycle mapping workflow. It is worth highlighting that the proposed ParaPoint, as an end-to-end unsupervised neural learning framework, shows obvious advantages in terms of flexibility and usability, compared with traditional mesh parameterization algorithms that require multiple cumbersome intermediate processing procedures and are difficult to implement and tune.

\begin{figure*}[t!]
	\centering
	\includegraphics[width=0.99\linewidth]{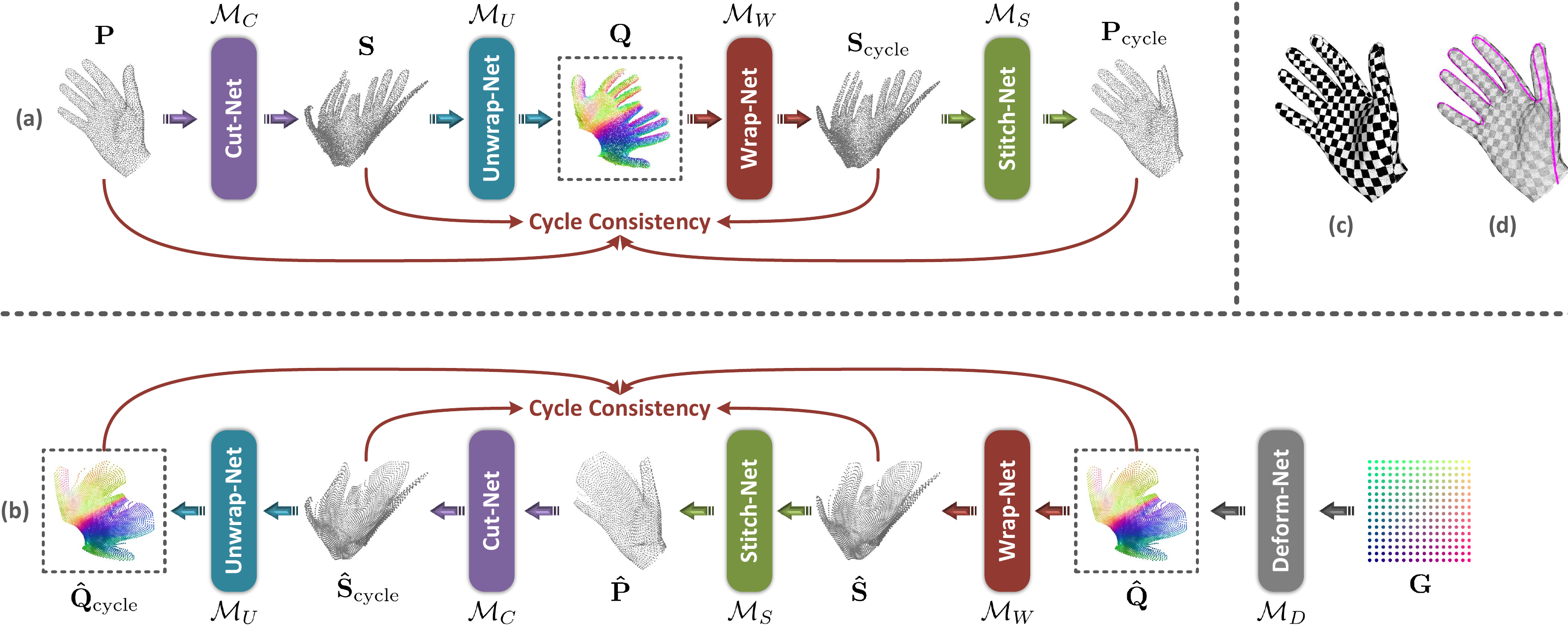}
	\caption{Illustration of the bi-directional cycle mapping pipeline, which is composed of (a) the 3D$\rightarrow$2D$\rightarrow$3D cycle mapping branch, and (b) the 2D$\rightarrow$3D$\rightarrow$2D cycle mapping branch. The input is an unoriented 3D point cloud containing $4096$ points. (c) visualizes the corresponding texture mapping. (d) shows the \textit{learned} cutting seams. The colors of 2D UV coordinates $\mathbf{Q}$, $\hat{\mathbf{Q}}$, and $\hat{\mathbf{Q}}_\mathrm{cycle}$ represent point-wise normal directions.}
	\label{fig:overall_workflow}
	\vspace{-0.3cm}
\end{figure*}

\textbf{Remarks.} Compared to the established techniques in mesh parameterization (refer to Section~\ref{sec:rw-mesh-para}), our neural parameterization framework introduces two distinct advantages: 
\begin{itemize}
	\item \textbf{Flexibility with input.} Traditional mesh parameterization methods typically necessitate input meshes to have relatively uniform triangulations to circumvent any numerical issues for the linear system solver or the optimization process of minimizing distortion-related energies. When confronted with meshes characterized by irregular triangulation, a remeshing preprocessing step becomes indispensable. In sharp contrast, our approach does not rely on any form of connectivity, allowing it to be directly applied to raw point clouds. This capability significantly reduces the complexity involved in converting point clouds into meshes, streamlining the parameterization process.
	\item \textbf{Smoothness of parameterization.} Mesh-based parameterization techniques are limited to computing parameters solely for the mesh vertices. Consequently, for any arbitrary point on the mesh that is not a vertex, these methods resort to bi-linear interpolation to determine the parameters $(u,v)$, utilizing the parameters of the vertices of the encompassing triangle. This approach, however, does not guarantee smoothness, particularly across edges, and this lack of smoothness is exacerbated in low-resolution meshes. Our method, leveraging the inherent smooth properties of neural networks, learns the parameterization in a manner that ensures smoothness for all points on the target surface, thereby overcoming the limitations of traditional mesh-based approaches.
	
\end{itemize}

\subsection{Network Structures} \label{sec:network-structures}

As illustrated in Fig.~\ref{fig:overall_workflow}, there are five sub-networks with different functionalities for manipulating both the 3D point cloud shape and the 2D UV points, as listed below:
\begin{itemize}
	
	\item The 2D boundary deforming network (\textbf{Deform-Net}), $\mathcal{M}_D: \mathbb{R}^2 \rightarrow \mathbb{R}^2$, takes a pre-defined 2D grid as input and adaptively deforms the initial grid points to potentially optimal 2D UV coordinates.
	
	\item The 3D surface cutting network (\textbf{Cut-Net}), $\mathcal{M}_C: \mathbb{R}^3 \rightarrow \mathbb{R}^3$, cuts the given shape surface to an open manifold in the 3D space that is developable.
	
	\item The 3D surface stitching network (\textbf{Stitch-Net}), $\mathcal{M}_S: \mathbb{R}^3 \rightarrow \mathbb{R}^3$, consumes a pre-cut open surface as input and stitches the seams to recover the surface of the given shape.
	
	\item The 2D-to-3D wrapping network (\textbf{Wrap-Net}), $\mathcal{M}_W: \mathbb{R}^2 \rightarrow \mathbb{R}^3$, smoothly maps the 2D UV coordinates to the 3D space. 
	
	\item The 3D-to-2D unwrapping network (\textbf{Unwrap-Net}), $\mathcal{M}_U: \mathbb{R}^3 \rightarrow \mathbb{R}^2$, flattens 3D points onto the 2D planar domain.
	
\end{itemize}

To avoid cumbersome notations used for describing the working mechanisms of these different sub-networks, below we uniformly use $\mathbf{X}^\mathrm{2D}$ and $\mathbf{X}^\mathrm{3D}$ to denote 2D and 3D point sets, and then indicate the network input or output in the right subscript. Note that all sub-networks are built upon point-wisely shared multi-layer perceptrons (MLPs) for implementing point-to-point non-linear mappings between input and output dimensions. For convenience, we use ${\bf\Phi}_{\ast \rightarrow \ast}(\cdot)$ to denote a stack of multiple MLPs, with the input and output channels indicated on the left and right side of the ``$\rightarrow$'' symbol, respectively. \\

\noindent \textbf{Deform-Net ($\mathcal{M}_D$) for 2D Boundary Deforming.} To achieve adaptive parameter domain deformation, we learn point-wise offsets from the initial 2D coordinates to produce a new set of UV points. Formally, the behavior of the Deform-Net can be formulated as:
\begin{equation}
	\mathbf{X}^\mathrm{2D}_\mathrm{out} = {\bf\Phi}_{(d+2) \rightarrow 2}([{\bf\Phi}_{2 \rightarrow d}(\mathbf{X}^\mathrm{2D}_\mathrm{in}); \mathbf{X}^\mathrm{2D}_\mathrm{in}]) + \mathbf{X}^\mathrm{2D}_\mathrm{in},
\end{equation}
\noindent where $[\cdot;\cdot]$ represents channel-wise concatenation. Here, the input 2D point set is first embedded through ${\bf\Phi}_{2 \rightarrow d}(\cdot)$  into the $d$-dimensional latent space, concatenated with itself, and then embedded back onto the 2D space through ${\bf\Phi}_{(d+1) \rightarrow 2}(\cdot)$. The resulting point-wise offsets are further added to the input 2D points to output the deformed UV points. \\

\noindent \textbf{Cut-Net ($\mathcal{M}_C$) for 3D Surface Cutting.} Flattening open surfaces with disk topology onto 2D planar domains is typically straightforward. Therefore, we propose to apply the Cut-Net to open the original shape to a highly developable surface manifold. The working mechanism of the cutting operation is basically the same as the Deform-Net, i.e., an offset-based learning process as given by: 
\begin{equation}
	\mathbf{X}^\mathrm{3D}_\mathrm{out} = {\bf\Phi}_{(d+3) \rightarrow 3}([{\bf\Phi}_{3 \rightarrow d}(\mathbf{X}^\mathrm{3D}_\mathrm{in}); \mathbf{X}^\mathrm{3D}_\mathrm{in}]) + \mathbf{X}^\mathrm{3D}_\mathrm{in},
\end{equation}
\noindent where we first produce point-wise 3D offsets through ${\bf\Phi}_{3 \rightarrow d}(\cdot)$ and ${\bf\Phi}_{(d+3) \rightarrow 3}(\cdot)$, and then add the offsets to the input 3D points to output the cutting results. \\

\noindent \textbf{Stitch-Net ($\mathcal{M}_S$) for 3D Surface Stitching.} The functionality of the Stitch-Net is opposite to the cutting effects of the Cut-Net, which can be formulated as:
\begin{equation}
	\mathbf{X}^\mathrm{3D}_\mathrm{out} = {\bf\Phi}_{3 \rightarrow 3}(\mathbf{X}^\mathrm{3D}_\mathrm{in}),
\end{equation}
\noindent where we directly transform the input pre-cut 3D surface manifold to recover the original shape geometry. Different from the Cut-Net that applies offset-based transformation, here we simply use a stack of MLPs to stitch the 3D points. \\

\noindent \textbf{Wrap-Net ($\mathcal{M}_W$) for 2D-to-3D Wrapping.} The Wrap-Net consumes adaptively deformed 2D UV points as input, and employs a stack of MLPs to generate an open 3D surface manifold, as given by:
\begin{equation}
	\mathbf{X}^\mathrm{3D}_\mathrm{out} = {\bf\Phi}_{2 \rightarrow 3}(\mathbf{X}^\mathrm{2D}_\mathrm{in}).
\end{equation}

\noindent \textbf{Unwrap-Net ($\mathcal{M}_U$) for 3D-to-2D Unwrapping.} The functionality of the Unwrap-Net is opposite to the wrapping effects of the Wrap-Net. The network input is supposed to be a highly developable 3D surface. Then a stack of MLPs is applied to map input 3D points onto the 2D parameter domain, as given by:
\begin{equation}
	\mathbf{X}^\mathrm{2D}_\mathrm{out} = {\bf\Phi}_{3 \rightarrow 2}(\mathbf{X}^\mathrm{3D}_\mathrm{in})
\end{equation}

\subsection{Bi-directional Cycle Mapping} \label{sec:method-bi-cy}

We assemble the aforementioned sub-networks to construct a bi-directional cycle mapping pipeline for learning a bijective mapping between 3D surface points and adaptively deformed 2D UV points. Note that the same sub-network is shared in both branches.

Specifically, there are two cycle mapping branches with opposite directions. As illustrated in Fig.~\ref{fig:overall_workflow}, the upper branch consumes the input 3D point cloud $\mathbf{P}$ to conduct the 3D$\rightarrow$2D$\rightarrow$3D cycle mapping, while the bottom branch starts from a pre-defined UV space to inversely conduct the 2D$\rightarrow$3D$\rightarrow$2D cycle mapping. The detailed learning procedures are presented in the following. \\

\noindent \textbf{3D$\rightarrow$2D$\rightarrow$3D Cycle Mapping Branch.} We begin with feeding the given 3D point cloud $\mathbf{P}$ into the Cut-Net to obtain an open surface manifold $\mathbf{S} \in \mathbb{R}^{N \times 3}$, which is flattened through the Unwrap-Net to produce 2D UV coordinates $\mathbf{Q}$. To form the cycle mapping, we further feed $\mathbf{Q}$ into the Wrap-Net to generate $\mathbf{S}_\mathrm{cycle} \in \mathbb{R}^{N \times 3}$, which is supposed to be an open surface manifold point-wisely equal to $\mathbf{S}$. And then the Stitch-Net is applied on $\mathbf{S}_\mathrm{cycle}$ to produce $\mathbf{P}_\mathrm{cycle}$, which is supposed to be point-wisely equal to $\mathbf{P}$ for recovering the original shape geometry. Formally, these learning procedures can be sequentially written as:
\begin{equation}
	\begin{aligned}
		&\mathbf{S}  = \mathcal{M}_C(\mathbf{P}),~~ \mathbf{Q}  = \mathcal{M}_U(\mathbf{S}); \\
		&\mathbf{S}_\mathrm{cycle}  = \mathcal{M}_W(\mathbf{Q}),~~ \mathbf{P}_\mathrm{cycle}  = \mathcal{M}_S(\mathbf{S}_\mathrm{cycle}).
	\end{aligned}
\end{equation}

\noindent \textbf{2D$\rightarrow$3D$\rightarrow$2D Cycle Mapping Branch.} We start by pre-defining a 2D lattice grid $\mathbf{G} \in \mathbb{R}^{N \times 2}$ uniformly sampled within $[-1, 1]^2$. Note that in practice the number of grid points in $\mathbf{G}$ is not strictly required to be the same as the number of input 3D points in $\mathbf{P}$, but here for simplicity, we assume that $N$ grid points are sampled. To perform adaptive deformation of 2D UV space, we feed $\mathbf{G}$ into the Deform-Net to produce a new set of 2D UV coordinates $\mathbf{\hat{Q}} \in \mathbb{R}^{N \times 2}$ as:
\begin{equation}
	\mathbf{\hat{Q}} = \mathcal{M}_D(\mathbf{G}).
\end{equation}

After that, we feed $\mathbf{\hat{Q}}$ into the Wrap-Net to obtain $\mathbf{\hat{S}} \in \mathbb{R}^{N \times 3}$, which is further stitched by the Stitch-Net to produce $\mathbf{\hat{P}}$ for recovering the original shape geometry. To form the cycle mapping, $\mathbf{\hat{P}}$ is opened by the Cut-Net to output $\mathbf{\hat{S}}_\mathrm{cycle}$, and then $\mathbf{\hat{S}}_\mathrm{cycle} \in \mathbb{R}^{N \times 3}$ is flattened by the Unwrap-Net to output $\mathbf{\hat{Q}}_\mathrm{cycle} \in \mathbb{R}^{N \times 2}$. Similarly, $\mathbf{\hat{Q}}$ and $\mathbf{\hat{S}}$ are expected to be point-wisely equal to $\mathbf{\hat{Q}}_\mathrm{cycle}$ and $\mathbf{\hat{S}}_\mathrm{cycle}$. Formally, these learning procedures can be sequentially written as:
\begin{equation}
	\begin{aligned}
		&\mathbf{\hat{S}} = \mathcal{M}_W(\mathbf{\hat{Q}}),~~ \mathbf{\hat{P}} = \mathcal{M}_S(\mathbf{\hat{S}}); \\
		&\mathbf{\hat{S}}_\mathrm{cycle} = \mathcal{M}_C(\mathbf{\hat{P}}),~~ \mathbf{\hat{Q}}_\mathrm{cycle} = \mathcal{M}_U(\mathbf{\hat{S}}_\mathrm{cycle}).
	\end{aligned}
\end{equation}

\noindent \textbf{Remark.} For a given 3D point cloud $\mathbf{P}$ fed into the bi-directional cycle mapping pipeline, we determine $\mathbf{Q}$ as the resulting point-wisely parameterized (i.e., row-wisely mapped) 2D UV coordinates. All the other intermediate results just serve for the overall optimization process. 

Moreover, after the training of ParaPoint, we are able to extract the cutting seams $\mathbf{E} = \{ \mathbf{e}_i \in \mathbf{P} \}$ by comparing $\mathbf{P}$ and $\mathbf{Q}$. Specifically, for each 3D point $\mathbf{p}_i \in \mathbf{P}$ and its K-nearest neighbors $\{ \mathbf{p}_i^{(k)} \}_{k=1}^{K_\mathrm{cut}}$, we know their 2D UV coordinates denoted as $\mathbf{q}_i \in \mathbf{Q}$ and $\{ \mathbf{q}_i^{(k)} \}_{k=1}^{K_\mathrm{cut}}$, respectively. Then we compute the maximum distance between $\mathbf{q}_i$ and its neighboring points as:
\begin{equation} \label{eqn:find-cut}
	d_i = \mathrm{max}(\{ \lVert \mathbf{q}_i - \mathbf{q}_i^{(k)} \rVert_2 \}_{k=1}^{K_\mathrm{cut}}).
\end{equation}
Thus, $\mathbf{p}_i$ can be determined to locate on the cutting seam when $d_i$ is larger than a threshold $T_\mathrm{cut}$.

\subsection{Training Objectives} \label{sec:training-objectives}

As an unsupervised learning pipeline, we train ParaPoint by minimizing a series of carefully designed loss functions and optimization constraints, whose detailed formulations and functionalities are introduced as follows. \\

\noindent \textbf{Unwrapping Loss.} We focus on the unwrapping process from input 3D points $\mathbf{P}$ to their corresponding 2D UV points $\mathbf{Q}$. The most fundamental requirement for $\mathbf{Q}$ is that any two parameterized coordinates cannot overlap. For each point $\mathbf{q}_i$, we search its K-nearest-neighbors $\{ \mathbf{q}_{i}^{(k)} \}_{k=1}^{K_u}$, and optimize the unwrapping loss to penalize points that are too close:
\begin{equation} \label{eqn:unwrap-loss}
	\ell_\mathrm{unwrap} = \sum_{i=1}^{N} \sum_{k=1}^{K_\mathrm{u}} \mathrm{max}(0, \epsilon - \lVert \mathbf{q}_i - \mathbf{q}_{i}^{(k)} \rVert_2),
\end{equation}
\noindent where the minimal distance allowed between neighboring points is controlled by a threshold $\epsilon$. \\

\noindent \textbf{Wrapping Loss.} We focus on the wrapping process from adaptively deformed 2D UV points $\hat{\mathbf{Q}}$ to 3D points $\hat{\mathbf{P}}$, which should approximate the original 3D shape geometry of $\mathbf{P}$. Since $\hat{\mathbf{P}}$ and $\mathbf{P}$ are not row-wisely corresponded, we formulate the wrapping loss as computing the similarity between the two 3D point sets:
\begin{equation}
	\ell_\mathrm{wrap} = \texttt{CD}(\hat{\mathbf{P}},~\mathbf{P}),
\end{equation}
where $\texttt{CD}(\cdot,~\cdot)$ represents the commonly-used Chamfer distance. \\

\noindent \textbf{Cycle Consistency Loss.} In our bi-directional cycle mapping framework, we seek to promote consistencies within both 3D and 2D domains, which are natural to learn thanks to our symmetric designs of the functionalities of sub-networks, i.e., (1) Cut-Net is inverse of Stitch-Net; and (2) Wrap-Net is inverse of Unwrap-Net. Thus, the cycle consistency loss is formulated as follows:
\begin{equation}
	\ell_\mathrm{cycle} = \lVert \mathbf{P} - \mathbf{P}_\mathrm{cycle} \rVert_1 + 
	\lVert \mathbf{S} - \mathbf{S}_\mathrm{cycle} \rVert_1 + 
	\lVert \hat{\mathbf{Q}} - \hat{\mathbf{Q}}_\mathrm{cycle} \rVert_1 + 
	\lVert \hat{\mathbf{S}} - \hat{\mathbf{S}}_\mathrm{cycle} \rVert_1
\end{equation}

\vspace{0.2cm}
\noindent \textbf{Remark.} We empirically observe that optimizing the above three loss functions $\ell_\mathrm{unwrap}$, $\ell_\mathrm{wrap}$, and $\ell_\mathrm{cycle}$, is already sufficient to produce reasonable parameterization results. Specifically, the original 3D shape surface is cut along clear seams, and the opened surface is then smoothly flattened onto 2D UV space, which is adaptively deformed. However, pursuing high-quality surface parameterization requires more precise and fine-grained constraints for the learned neural mapping. Therefore, we further impose the following constraints.

\vspace{1em}
\noindent \textbf{Mapping Distortion Loss.} We regularize the parameterization distortions of the learned neural mapping function by exploiting the differential surface properties, \textit{which can be conveniently deduced via the automatic differentiation mechanisms in common deep learning programming frameworks}.

In our bi-directional cycle mapping pipeline, we have the 2D UV coordinates $\mathbf{Q}$ and $\hat{\mathbf{Q}}$ mapped to the 3D surface points $\mathbf{P}_\mathrm{cycle}$ and $\hat{\mathbf{P}}$, respectively. We denote the corresponding neural mapping functions $f: \mathbb{R}^2 \rightarrow \mathbb{R}^3$ and $g: \mathbb{R}^2 \rightarrow \mathbb{R}^3$:
\begin{equation}
	\begin{aligned}
		\mathbf{P}_\mathrm{cycle} = \:&f(\mathbf{Q}) = \mathcal{M}_S(\mathcal{M}_W(\mathbf{Q})); \\
		\hat{\mathbf{P}} = \:&g(\hat{\mathbf{Q}}) = \mathcal{M}_S(\mathcal{M}_W(\hat{\mathbf{Q}})).
	\end{aligned}
\end{equation}

We compute the derivatives of $f$ and $g$ with respect to $\mathbf{Q}$ and $\hat{\mathbf{Q}}$ at each 2D UV point $(u, v)$ to obtain the Jacobian matrices $\mathbf{J}_f \in \mathbb{R}^{3 \times 2}$ and $\mathbf{J}_g \in \mathbb{R}^{3 \times 2}$:
\begin{equation}
	\mathbf{J}_f = (f_u \enspace f_v), ~ \mathbf{J}_g = (g_u \enspace g_v)
\end{equation}
\noindent where $f_u$, $f_v$, $g_u$, $g_v$ are three-dimensional vectors of partial derivatives. After these, we further compute the eigenvalues of $\mathbf{J}_f$ and $\mathbf{J}_g$, denoted as $(\lambda_f^1, \lambda_f^2)$ and $(\lambda_g^1, \lambda_g^2)$. Thus, we can promote conformal (angle-preserving) parameterizations by constraining the following regularizer:
\begin{equation}
	\ell_\mathrm{conf} = \\
	\sum_{\mathbf{q} \in \mathbf{Q}} \lVert \lambda_f^1 - \lambda_f^2 \rVert_1 + \\ 
	\sum_{\hat{\mathbf{q}} \in \hat{\mathbf{Q}}}\lVert \lambda_g^1 - \lambda_g^2 \rVert_1.
\end{equation}

\vspace{-0.1cm}
Alternatively, we can choose to promote isometric parameterizations by formulating a stronger regularization term $\ell_\mathrm{isom}$ that requires each eigenvalue equal to $1$ by simply measuring $L_1$ distances. \textit{Still, unless particularly stated, our subsequent experiments default to only applying the conformal constraint $\ell_\mathrm{conf}$.} \\

\noindent \textbf{Anti-Flipping Loss.} Having obtained $(f_u \enspace f_v)$ and $(g_u \enspace g_v)$, it is straightforward to compute the normal vectors $\mathbf{n}_f \in \mathbb{R}^3$ and $\mathbf{n}_g\in \mathbb{R}^3$ by:
\begin{equation}
	\mathbf{n}_f = \frac{ f_u \times f_v }{ \| f_u \times f_v \| }, ~ \mathbf{n}_g = \frac{ g_u \times g_v }{ \| g_u \times g_v \| }.
\end{equation}

In this way, for each 3D surface point $\mathbf{p}_\mathrm{cycle} \in \mathbf{P}_\mathrm{cycle}$ and $\hat{\mathbf{p}} \in \hat{\mathbf{P}}$, we deduce their normal vectors $\mathbf{n}_\mathrm{cycle}$ and $\hat{\mathbf{n}}$ in a differentiable manner. To promote flip-free parameterizations, we formulate the anti-flipping loss by penalizing neighboring points whose normal directions are significantly different:
\begin{equation} \label{eqn:aflip-loss}
	\begin{aligned}
		\ell_\mathrm{aflip} = &\sum\nolimits_{\mathbf{p}_\mathrm{cycle}} \sum\nolimits_{\mathbf{p}_\mathrm{cycle}^{(k)}} \mathrm{max}(0, \, \texttt{AD}(\mathbf{n}_\mathrm{cycle} ; \mathbf{n}_\mathrm{cycle}^{(k)}) - T_\mathrm{angle}) \\
		+ &\sum\nolimits_{\hat{\mathbf{p}}} \sum\nolimits_{\hat{\mathbf{p}}^{(k)}} \mathrm{max}(0, \, \texttt{AD}(\hat{\mathbf{n}} ; \hat{\mathbf{n}}^{(k)}) - T_\mathrm{angle})
	\end{aligned}
\end{equation}
\noindent where $\mathbf{p}_\mathrm{cycle}^{(k)}$ and $\hat{\mathbf{p}}^{(k)}$, whose normal vectors are $\mathbf{n}_\mathrm{cycle}^{(k)}$ and $\hat{\mathbf{n}}^{(k)}$, are within the K-nearest-neighbors of $\mathbf{p}_\mathrm{cycle}$ and $\hat{\mathbf{p}}$. Here, $\texttt{AD}(\cdot;\cdot)$ measures the cosine similarity, such that angle differences are expected not to exceed a threshold $T_\mathrm{angle}$. \\

\begin{figure*}[h!]
	\centering
	\vspace{-0.6cm}
	\includegraphics[width=0.99\linewidth]{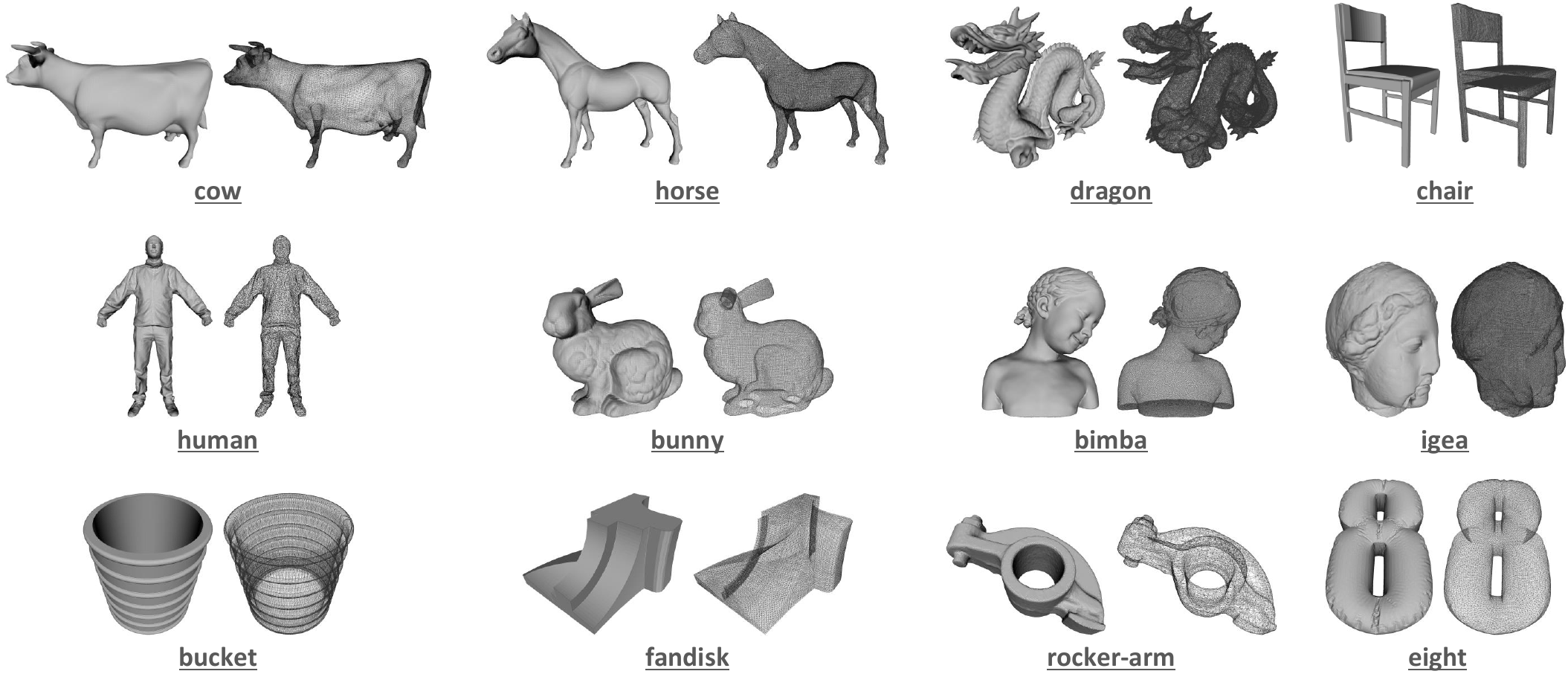}
	\caption{Display of various selected testing shapes, where we also present the mesh edges to facilitate showing the shape structure. Note that the inputs to our approach are only 3D point coordinates (without normals) sampled from the target surfaces.}
	\label{fig:testing_models_display}
	\vspace{-0.8cm}
\end{figure*}

\section{Experiments} \label{sec:experiments}

\subsection{Implementation Details} \label{sec:exp-impl}

In our bi-directional cycle mapping pipeline, all sub-networks are architecturally built upon stacked MLPs without batch normalization. We uniformly configured LeakyReLU non-linearity with the negative slope of $0.01$, except for the output layer. For the Wrap-Net, Unwrap-Net, and Stitch-Net consisting of one stack of $5$ MLPs, the number of feature channels in the hidden layers are sequentially set as $[64, 128, 512, 128]$. For the Deform-Net and Cut-Net composed of two stacks of $5$ MLPs, we configured $[64, 128, 512, 128]$ and $d=64$ for hidden space embedding, and $[64, 128, 512, 128]$ for offset generation.

\begin{figure*}[t!]
	\centering
	\includegraphics[width=0.99\linewidth]{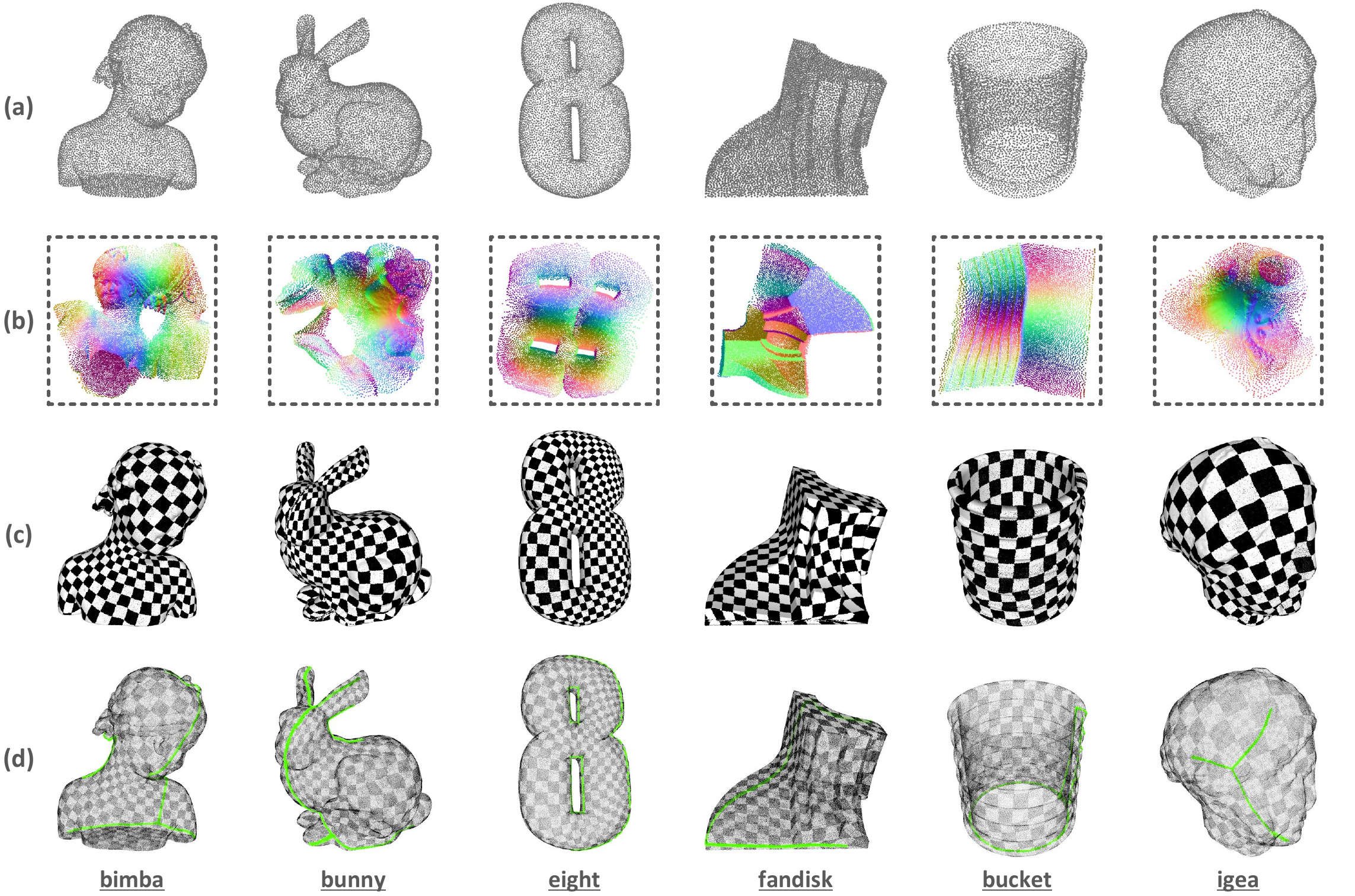}
	\caption{Illustration of neural parameterization results produced by ParaPoint. (a) Input 3D point cloud; (b) 2D UV coordinates; (c) Texture mapping from the checkerboard; (d) The automatically learned cutting seams (points marked in \textcolor{green}{green}). }
	\label{fig:para_results_display_1}
	\vspace{-0.3cm}
\end{figure*}

In addition to network structures, there is also a set of hyperparameters to be configured and tuned. As presented in Eqn.~(\ref{eqn:find-cut}) for cutting seam extraction, we set $K_\mathrm{cut} = 3$. Suppose that, at the current training iteration, the side length of a square bounding box of 2D UV coordinates $\mathbf{Q}$ is denoted as $L(\mathbf{Q})$. Then we set the threshold $T_\mathrm{cut}$ to be $1\%$ of $L(\mathbf{Q})$. Besides, as presented in Eqn.~(\ref{eqn:unwrap-loss}) for computing the unwrapping loss, we choose the threshold as $\epsilon = 0.1 \cdot L(\mathbf{Q}) / \sqrt{N}$. As presented in Eqn.~(\ref{eqn:aflip-loss}) for computing anti-flipping loss, we set $T_\mathrm{angle} = \pi/2$, and the neighborhood size is chosen as $4$. When formulating the overall training objective, the weights for $\ell_\mathrm{unwrap}$, $\ell_\mathrm{wrap}$, $\ell_\mathrm{cycle}$, and $\ell_\mathrm{aflip}$ are fixed to $0.01$, $1.0$, $0.01$, and $0.01$, respectively. For constraining mapping distortions, we empirically found that the weight for $\ell_\mathrm{conf}$ should be weakened when dealing with complex geometric and topological structures, but $0.01$ still works well on many common 3D shapes. Besides, for some particular types of surfaces (e.g., cube, cylinder), replacing $\ell_\mathrm{conf}$ with $\ell_\mathrm{isom}$ should achieve the optimal results.

Instead of randomly initializing neural network parameters, we conducted a warming-up procedure for the input 3D shape via pre-fitting our framework on its convex hull. During the optimization stage, we started by feeding a sparser subset of the original point cloud to speed up the whole learning process, and feed all the given points in later training iterations. We also slightly perturbed the point coordinates to promote the continuity and robustness of the learned neural mapping.

\begin{figure*}[t!]
	\centering
	\includegraphics[width=0.99\linewidth]{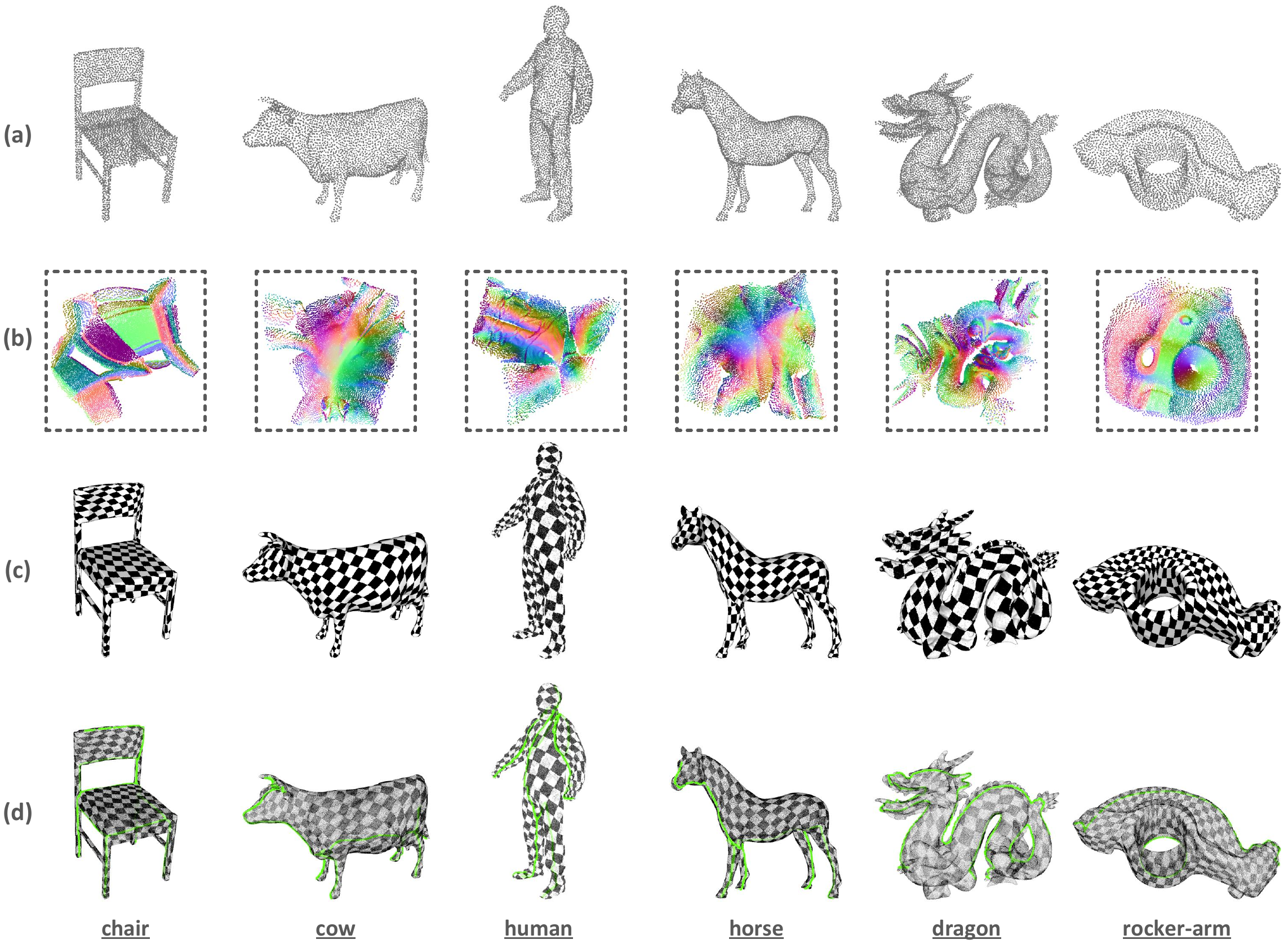}
	\caption{Neural parameterization for complex 3D shapes, where the weight for the $\ell_\mathrm{conf}$ is decreased from the default $1e^{-2}$ to $1e^{-4}$ for relaxing the conformal constraint.}
	\label{fig:para_results_display_2}
	\vspace{-0.3cm}
\end{figure*}

\subsection{UV Unwrapping and Texture Mapping} \label{sec:exp-applications}

We experimented with a variety of 3D shape models that are commonly used in the geometry processing community, covering different complexities of geometric and topological structures, as displayed in Fig.~\ref{fig:testing_models_display}.

We prepared the input point cloud data by sampling $10,000$ unoriented surface points from the testing 3D shape models. After training, we deduced point-wise UV mappings and performed color-coding using the ground-truth normals. After shown in  Figs.~\ref{fig:para_results_display_1} (\textcolor{red}{a}) and (\textcolor{red}{b}), ParaPoint can learn continuous and smooth 3D surface flattenings and adaptively deform 2D UV boundaries. Moreover, we directly input densely-sampled (around $250,000$) surface points without further fine-tuning the trained neural networks, and exploited the resulting UV unwrapping results to perform checkerboard texture mappings, achieving satisfactory visual effects as illustrated in Fig.~\ref{fig:para_results_display_1} (\textcolor{red}{c}). More surprisingly, as depicted in Fig.~\ref{fig:para_results_display_1} (\textcolor{red}{d}), our approach can automatically learn highly meaningful cutting seams. In Fig.~\ref{fig:para_results_display_2}, we presented our neural parameterization results over more challenging 3D shape models. It turns out that ParaPoint still produces highly satisfactory performances.

\subsection{Ablative Analyses} \label{sec:exp-ablation}

\begin{figure*}[t!]
	\centering
	\includegraphics[width=0.99\linewidth]{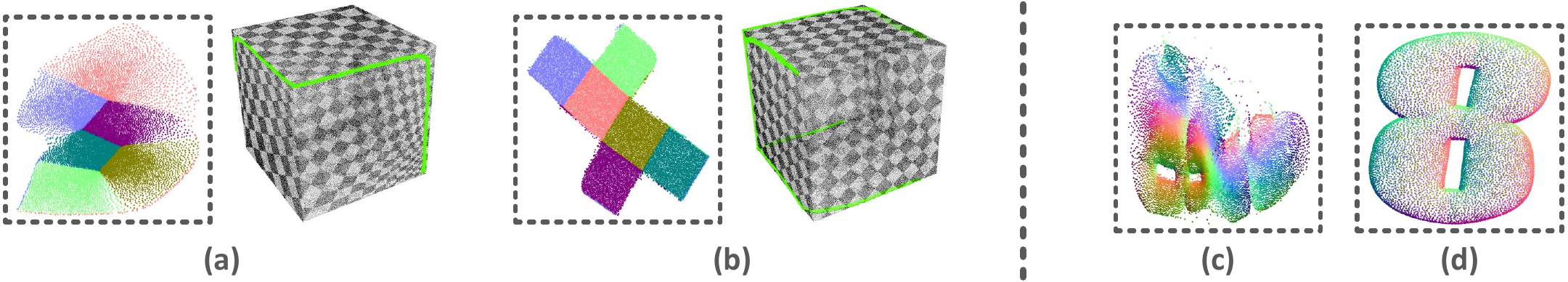}
	\caption{Ablation results in terms of distortion optimization objectives and bi-directional mapping branches. (a) and (b) compare the effects of constraining conformal and isometric losses. (c) and (d) show the UV unwrapping results of the ``eight'' model produced by only preserving the 2D$\rightarrow$3D$\rightarrow$2D branch and the 3D$\rightarrow$2D$\rightarrow$3D branch, respectively.}
	\label{fig:ablation_results}
	\vspace{-0.3cm}
\end{figure*}

As aforementioned, constraining conformality might not always be the optimal choice. For example, as shown in Figs.~\ref{fig:ablation_results} (\textcolor{red}{a}) and (\textcolor{red}{b}), optimizing the isometric loss leads to the most reasonable parameterization on a 3D cube.

Furthermore, we verified that simply applying uni-directional cycle mapping cannot produce high-quality surface parameterization results. As compared in Figs.~\ref{fig:ablation_results} (\textcolor{red}{c}) and (\textcolor{red}{d}), only training a 2D$\rightarrow$3D$\rightarrow$2D cycle mapping leads to highly distorted and cluttered distribution of UV coordinates, while only preserving the 3D$\rightarrow$2D$\rightarrow$3D branch totally fails to cut the original 3D surface to an open disk for subsequent flattening.

\section{Conclusion and Discussion} \label{sec:conclusion}

We presented ParaPoint, the \textit{first} neural parameterization pipeline operating on unstructured and unoriented 3D point clouds for achieving global free-boundary UV mapping. The core component of our approach is a highly interpretable bi-directional cycle mapping framework, which is composed of several geometrically meaningful sub-networks with specific functionalities, and trained with carefully designed loss functions and differential geometric constraints. We experimented with multiple 3D models with varying complexities of geometries and topologies, and found that the proposed ParaPoint is able to learn reasonable cutting seams and UV boundaries in a fully unsupervised manner.

We see many future directions for further improving the current methodology. Considering that we are already able to extract clear-cutting seams, it shouldn't be hard to pursue seamless surface parameterization by imposing constraints on the gradients of 2D UV points located on boundaries. Additionally, we wish to extend our approach to a generalizable neural architecture, instead of a per-input overfitting workflow. We believe that the proposed neural mapping paradigm shows great potential and brings novel insights that may open up many new possibilities for the field of surface parameterization.

\clearpage
\bibliographystyle{splncs04}

\newpage
\onecolumn

\centerline{\textbf{\textit{\Large (Supplementary Materials)}}}

\vspace{1.0cm}

Commonly-used traditional mesh parameterization algorithms \cite{levy2023least,sheffer2005abf++,DBLP:journals/tog/RabinovichPPS17} are only directly applicable to 3D surfaces homeomorphic to a disk, while our proposed ParaPoint is highly flexible to deal with complex topologies thanks to its ability to automatically perform shape cutting with reasonable seams. Therefore, in our paper, we did not make comparisons with these traditional approaches because they require additionally-specified cutting seams.

\begin{figure*}[h!]
	\centering
	\vspace{-0.6cm}
	\includegraphics[width=0.8\linewidth]{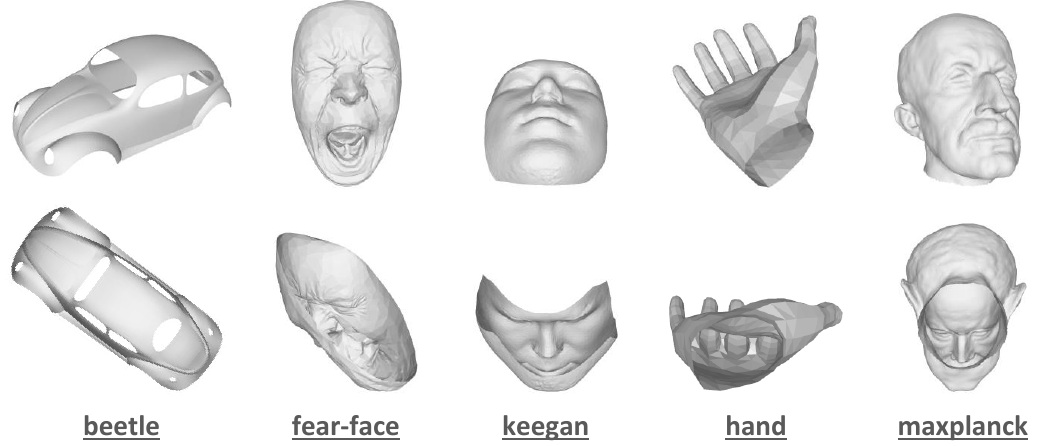}
	\caption{Display of our selected testing models with disk topologies, where each model is viewed from two different perspectives.}
	\label{suppl-fig:disk_topo_testing_models}
	\vspace{-1.3cm}
\end{figure*}

\begin{table}[h!]
	\centering	
	\renewcommand\arraystretch{1.25}
	\setlength{\tabcolsep}{8.0pt}
	\caption{Comparisons of parameterization conformality (i.e., angle-preserving) metrics (the lower, the better).}
	\vspace{-0.3cm}
	\begin{tabular}{ c | c | c | c | c | c } 
		\toprule[1.0pt]
		Model & \textit{beetle} & \textit{fear-face} & \textit{keegan} & \textit{hand} & \textit{maxplanck}  \\
		\hline\hline
		SLIM~\cite{DBLP:journals/tog/RabinovichPPS17} & 0.410 & 0.635 & 0.063 & 0.609 & 0.307 \\
		\hline
		ParaPoint & 0.066 & 0.120 & 0.059 & 0.182 & 0.117 \\
		\bottomrule[1.0pt]
	\end{tabular}
	\vspace{-0.6cm}
	\label{suppl-tab:comparison_angle_distortion}
\end{table}

\begin{figure*}[t!]
	\centering
	\includegraphics[width=0.95\linewidth]{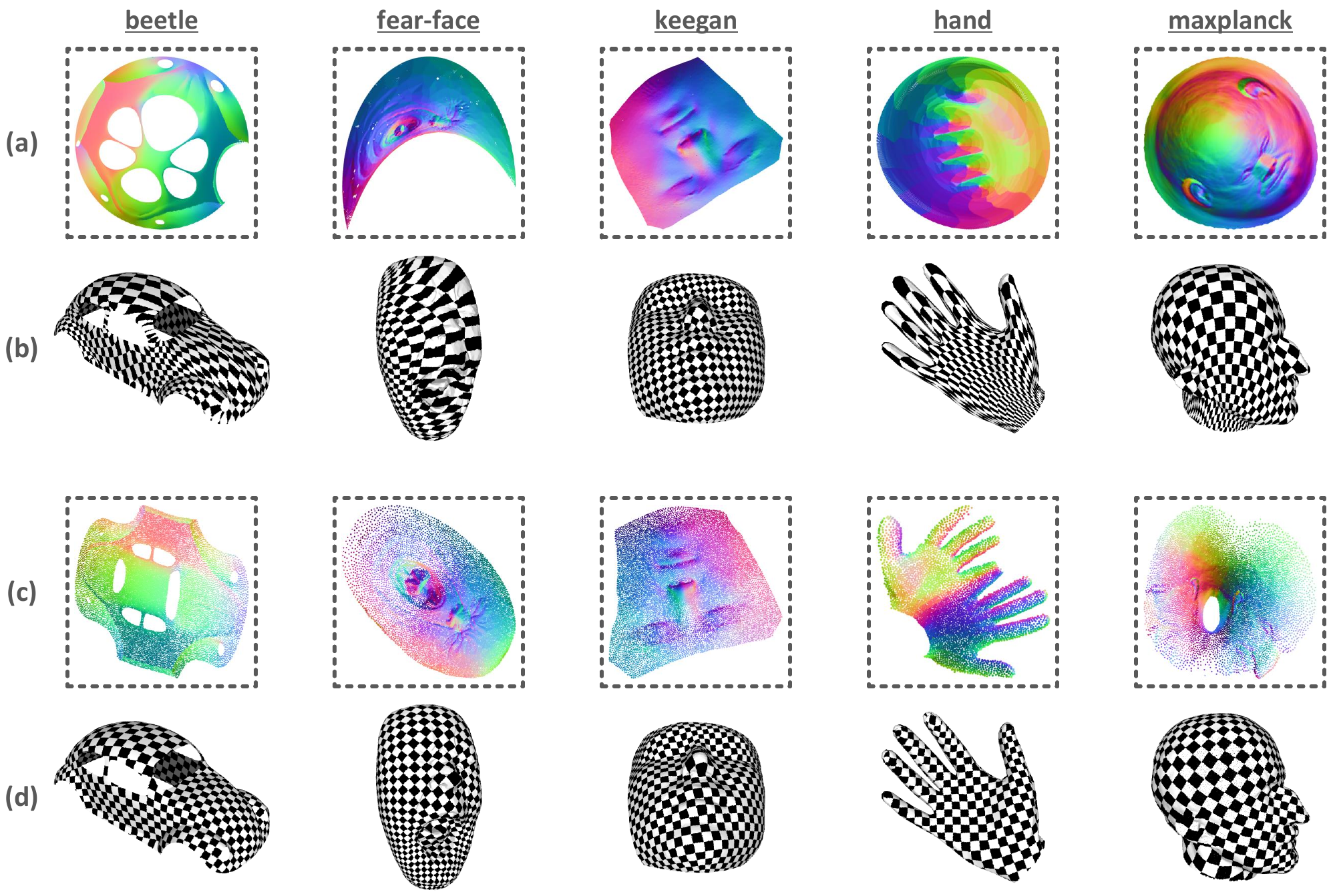}
	\caption{Display of testing models with disk topology. Each model is viewed from two different perspectives. (a) and (b) are produced by SLIM. (c) and (d) are our results.}
	\label{suppl-fig:disk_topo_para_results}
	\vspace{-0.3cm}
\end{figure*}

In this supplementary material, we attempted to make necessary qualitative and quantitative comparisons with one popular and advanced mesh parameterization algorithm SLIM~\cite{DBLP:journals/tog/RabinovichPPS17} by conducting experiments over testing models with disk topology, as displayed in Fig.~\ref{suppl-fig:disk_topo_testing_models}. We employed the officially released code of SLIM~\cite{DBLP:journals/tog/RabinovichPPS17} to parameterize the $5$ triangular mesh models. For our approach, we uniformly sample $10,000$ points from the original mesh vertices as inputs. After training, all the mesh vertices are fed into ParaPoint to produce per-vertex UV coordinates for performing checkerboard texture mappings as well as computing quantitative parameterization distortion metrics. Here, we evaluated conformality (i.e., angle preservation) by measuring the average of the absolute angle differences between each corresponding angle of the 3D triangles and the 2D parameterized triangles. Fig.~\ref{suppl-fig:disk_topo_para_results} displays the parameterization results produced by SLIM~\cite{DBLP:journals/tog/RabinovichPPS17} and our ParaPoint. The quantitative distortion metrics are reported in Tab.~\ref{suppl-tab:comparison_angle_distortion}. We can observe that our approach outperforms SLIM both qualitatively and quantitatively. Our adaptively deformed UV region is more reasonable, and the resulting texture mappings are more uniform and suffer from less stretches.

\begin{figure*}[t!]
	\centering
	\includegraphics[width=0.9\linewidth]{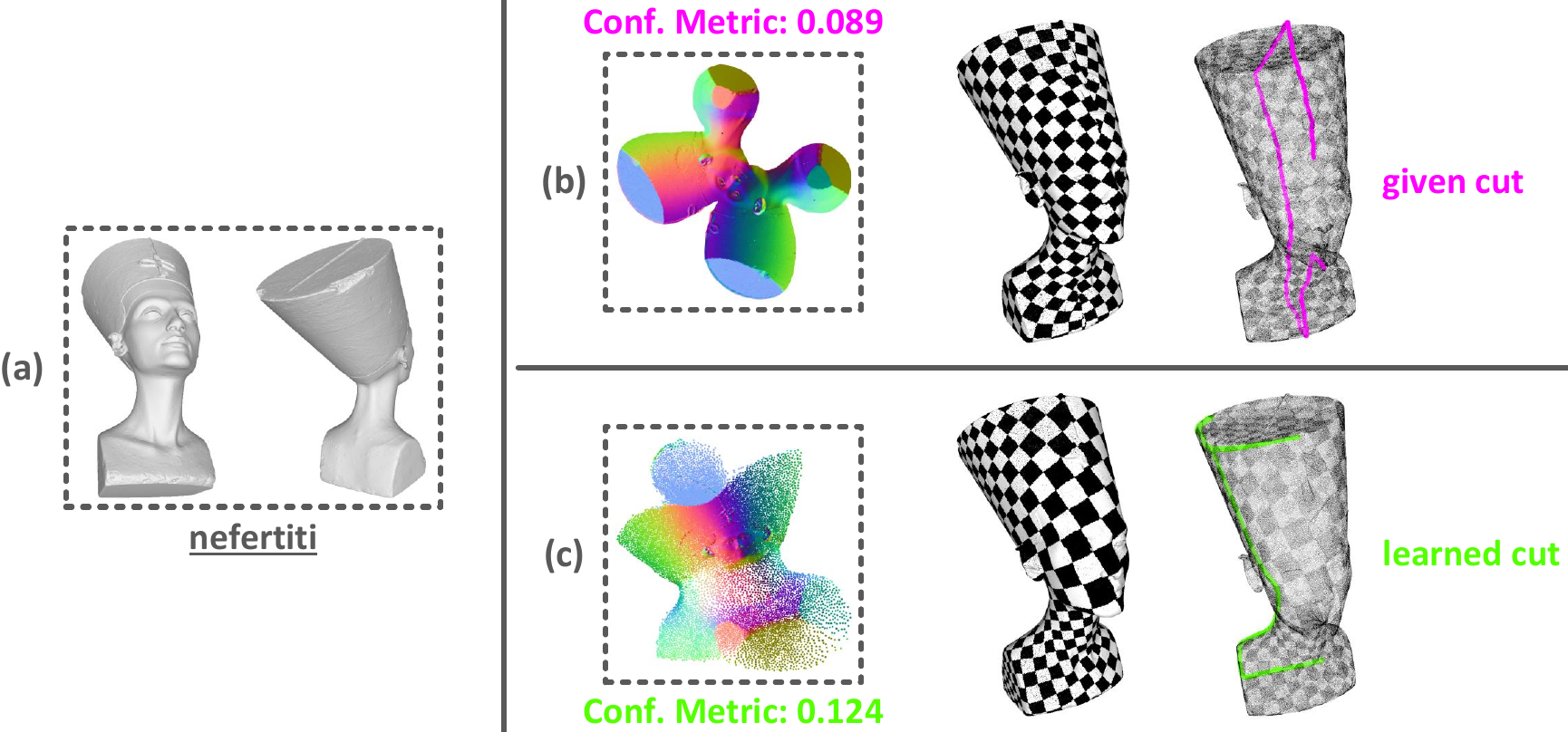}
	\caption{Experiments on the (a) \textit{nefertiti} testing model whose topology is NOT homeomorphic to a disk. (b) The parameterization results of SLIM are obtained by manually-specifying a high-quality cutting seam, with the conformality metric of $0.089$. (c) The parameterization results of our approach, in which the cutting seam is automatically learned, with the conformality metric of $0.124$.}
	\label{suppl-fig:compare_results_given_cut}
	\vspace{-0.6cm}
\end{figure*}

Moreover, to perform evaluations on shape models without disk topology, we made additional efforts to manually specify a potentially optimal cutting seam, such that SLIM can be applied. Note that this is a quite unfair experimental setting, because finding optimal cuts is known to be a critical yet complicated task. As shown in Fig.~\ref{suppl-fig:compare_results_given_cut}, our approach still achieves highly competitive performances compared with SLIM even under such an unfair comparison setup.

\end{document}